\pdfoutput=1
\documentclass[11pt]{article}

\usepackage[T1]{fontenc}
\usepackage[utf8]{inputenc}
\usepackage{amsmath,amssymb,amsthm}
\usepackage{graphicx}
\usepackage[table]{xcolor}
\usepackage{booktabs}
\usepackage{multirow}
\usepackage{array}
\usepackage{tabularx}
\usepackage{float}
\usepackage{algorithm}
\usepackage{algorithmic}
\usepackage{enumitem}
\usepackage{caption}
\usepackage{microtype}
\usepackage[margin=1in]{geometry}
\usepackage[numbers,sort&compress]{natbib}
\usepackage[colorlinks=true,linkcolor=blue,citecolor=blue,urlcolor=blue]{hyperref}
\usepackage[nameinlink]{cleveref}

\usepackage{amsmath,amsfonts,bm}

\def\eqref#1{equation~\ref{#1}}
\def\1{\bm{1}}

\DeclareMathAlphabet{\mathsfit}{\encodingdefault}{\sfdefault}{m}{sl}
\SetMathAlphabet{\mathsfit}{bold}{\encodingdefault}{\sfdefault}{bx}{n}

\def\sA{{\mathbb{A}}}

\def\sG{{\mathbb{G}}}

\def\sR{{\mathbb{R}}}
\def\sS{{\mathbb{S}}}

\newcommand{\E}{\mathbb{E}}

\newcommand{\R}{\mathbb{R}}

\title{Imagine to Ensure Safety in Hierarchical Reinforcement Learning}
\author{%
Gregory Gorbov$^{2,3}$, Artem Latyshev$^{1,2}$ and Aleksandr I. Panov$^{1,2}$\\[0.5em]
\small $^{1}$ Cognitive AI Systems Lab, Moscow 117312, Russia; \texttt{latyshev.a@miriai.org}; \texttt{panov@cogailab.com}\\
\small $^{2}$ Moscow Independent Research Institute of Artificial Intelligence, Moscow 141701, Russia; \texttt{gorbov.g@miriai.org}\\
\small $^{3}$ FRC Computer Science RAS, Moscow 119333, Russia\\
\small Correspondence: \texttt{panov@cogailab.com}
}
\date{}

\begin{document}
\maketitle

\begin{abstract}
This work investigates the safe exploration problem in reinforcement learning, where an agent must maximize cumulative performance while simultaneously satisfying safety constraints. This challenge becomes even more pronounced in long-horizon tasks, where existing safe methods face fundamental limitations due to compounding estimation errors and restricted exploration capabilities. To address this problem, we propose a method that combines a learnable world model with two complementary policies—a high-level policy and a low-level policy—to promote safety at both hierarchical levels. The high-level policy generates intermediate subgoals that bias exploration toward safe regions, while the low-level policy uses imagined rollouts in the learned world model to reduce unsafe behaviors when reaching these subgoals. The proposed method was evaluated on challenging long-horizon navigation and manipulation tasks with high-dimensional action spaces, where it significantly outperforms existing Safe RL baselines in both success rate and strong empirical constraint satisfaction, consistently meeting the prescribed safety budget across seeds, while prior approaches fail to effectively solve these complex long-horizon scenarios.
\end{abstract}


\section{Introduction}\label{sec:introduction}

Exploration is one of the most critical capabilities for Reinforcement Learning (RL) agents, enabling them to discover optimal behaviors for achieving predefined goals. One of the main challenges in applying RL algorithms to the real world is the lack of safety during the exploration process, which can lead to damage to expensive hardware or dangerous situations \cite{safetygym}. Therefore, there is an urgent need for Safe RL approaches that address the Safe Exploration problem, which we investigate in this work.

Conventional Safe RL approaches typically adopt the Constrained Markov Decision Process (CMDP) \cite{cmdpbook} framework, which incorporates an additional cost function to regulate action safety through constraint enforcement. Most current Safe RL methods optimize a composite objective function comprising two components -- reward and cost (e.g., Risk-Sensitive Criterion, Lagrangian-Based Approach, SafetyDreamer\cite{safedreamer}) \cite{garcia2015comprehensive, brunke2022safe, mbppol, safeslac, saclagrangian}. These approaches enable the RL controller to maximize performance while minimizing safety constraint violations. However, as demonstrated in \cite{safeditor}, these methods face a fundamental challenge: the inherent difficulty for the controller to simultaneously optimize performance and safety constraints.
Furthermore, these approaches typically employ flat-policy-based architectures (single-level policy), which fundamentally restrict their capability to solve long-horizon tasks --- a well-documented shortcoming evidenced in Hierarchical RL (HRL) literature \cite{chane2021goal, nachum2018data}. This architectural constraint consequently limits their practical applicability in complex, temporally extended scenarios where multi-level decision-making is essential. \textbf{Consequently, there exists a critical need for novel approaches capable of addressing long-horizon decision-making while rigorously maintaining safety constraints.}

The long-horizon problem is typically characterized by either: (1) an extended sequence of environment steps, (2) complex agent dynamics, or (3) sparse reward setting. Current solutions to this challenge predominantly involve demonstration learning, curriculum strategies, or HRL approaches \cite{levy2017hierarchical, hrac, mandlekar2020learning, bassich2019continuous}. Demonstration-based and curriculum approaches require substantial engineering effort and domain expertise, whereas HRL methods offer greater generalization across diverse tasks. 

The literature exhibits a significant gap in methods that effectively combine HRL with Safe RL approaches. Currently known techniques such as LyapunovRRT \cite{lyapunovrrt}, IAHRL \cite{iahrl}, and Safe HIRO \cite{safetylayerhiro} fail to address the challenge of learning in long-horizon environments while maintaining specified safety constraints in partial observability settings. These methods present substantial limitations: they require complete environment state information, rely on manually generated options that demand expert knowledge for each specific environment, lack the ability to generate safe subgoals needed to ensure safety throughout the learning process.

In this work, we propose a method \textbf{ITES} (\textbf{I}magine \textbf{T}o \textbf{E}nsure \textbf{S}afety in HRL) 
that uses a high-level policy to simplify the complex optimization task for the RL controller by generating \textbf{safe intermediate subgoals} (Figure~\ref{fig:conceptual_scheme}). We also use the world model to verify the safety of actions taken by the RL controller in the imagination (predicting next steps in the world model) before executing them directly. Furthermore, we identify a systematic limitation of existing state-of-the-art Safe RL approaches when deployed in high-dimensional, long-horizon robotic domains. While these methods often attain competitive results on short-horizon benchmarks, their performance deteriorates markedly as task complexity increases, leading to violations of safety constraints and reduced task completion. In contrast, ITES exhibits substantially improved robustness and success rates while consistently satisfying safety requirements, thereby effectively addressing scenarios in which prior methods fail to generalize.

Our main contributions are:
\begin{itemize}
    \item Our work identifies a fundamental limitation in current Safe RL methods employing flat-policy-based architectures: their inability to solve long-horizon tasks while preserving safety. To address this critical challenge, we propose ITES, a novel framework that successfully resolves this dual requirement.
    \item We propose a safe hierarchical method --- using a world model to ensure the safety of the controller, as well as a cost model for generating safe subgoals.
    \item We introduce a novel state-of-the-art approach for high-dimensional, long-horizon robotic tasks, demonstrating significant improvements in both safety and performance while maintaining the required safety level. Additionally, we show that on short-horizon tasks our method also outperforms existing Safe RL state-of-the-art approaches in terms of task success, again preserving the mandated safety constraints.
\end{itemize}

The implementation of the proposed method is publicly available at:
\url{https://github.com/Gricha1/ITES_alg}

\begin{figure}[H]
\includegraphics[width=0.7\textwidth]{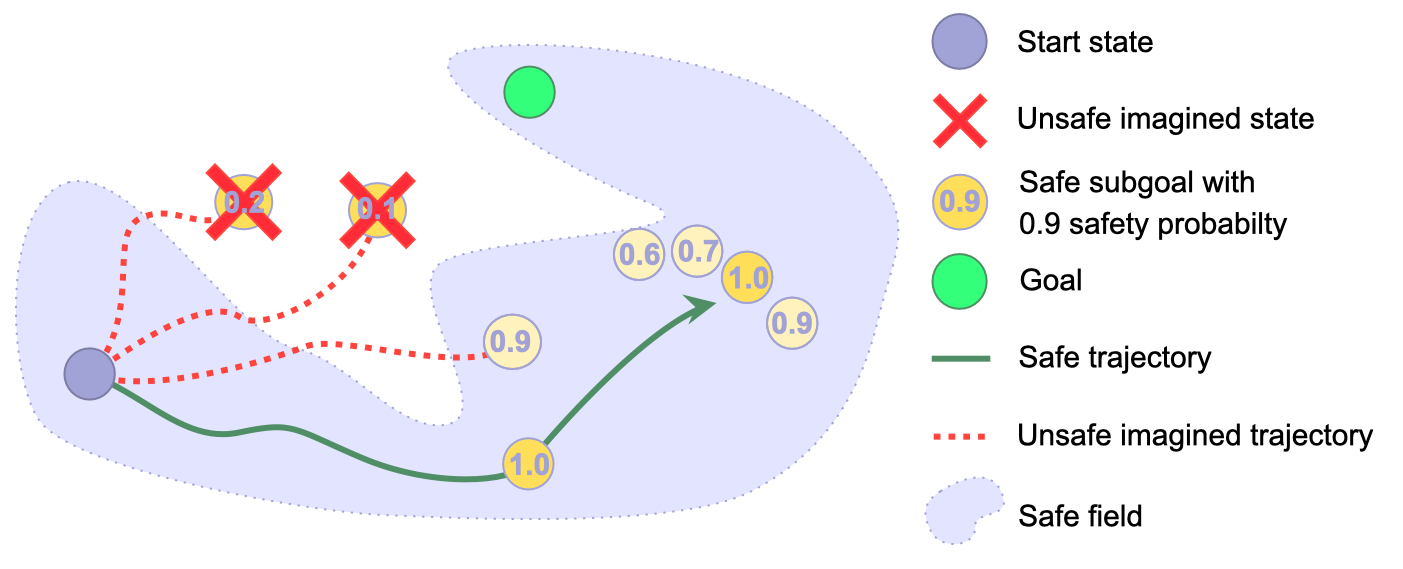}
\caption{The ITES generates intermediate subgoals, \textbf{orange circles}, to achieve the main goal, \textbf{green circle}, which enables the controller to optimize safety over a short-horizon for each of these subgoals. To ensure that the entire trajectory remains safe, the proposed method first predicts a safe subgoal for the controller, and second, the controller in imagination (via world model) optimizes safety towards that subgoal.\label{fig:conceptual_scheme}}
\end{figure}

\section{Related Work}\label{sec:related_work}

\textbf{Safe RL Methods}.
Modern Safe RL approaches can be broadly grouped into three categories: (1) stability-based methods, such as Lyapunov functions and control barrier functions, which require access to system dynamics; (2) intervention-based techniques (e.g., shielding), which rely on hand-crafted safety rules; and (3) constrained optimization methods, including Risk-Sensitive Criterion \cite{huang2021risk} and Lagrangian-based approaches such as TD3-Lagrangian \cite{saclagrangian}, CUP \cite{cup}, FOCOPS \cite{focops}, and SafeDreamer \cite{safedreamer}.
Notably, this third class of methods offers greater generality than both stability-based and intervention-based approaches, as it can be applied directly in partially observable environments without requiring either accurate system dynamics or manually specified safety rules.

CUP and FOCOPS approximate the cost constraint via a Taylor expansion to obtain tractable updates and theoretical safety guarantees; however, this approximation inherently introduces systematic estimation errors --- a well-known limitation that persists regardless of the task horizon. TD3-Lagrangian instead directly optimizes the dual Lagrangian formulation of the CMDP objective. SafeDreamer adapts Dreamer by incorporating Constrained CEM planning within a learned world model and currently achieves state-of-the-art performance on short-horizon SafetyGym tasks.

Despite their empirical success, constrained optimization methods face several fundamental challenges: dual optimization instability \cite{safeditor}, sample inefficiency, and insufficient exploration, all of which become particularly severe in sparse-reward settings or environments with large state and action spaces.

\textbf{HRL}. HRL frameworks (HAC \cite{levy2017hierarchical}, HiRO \cite{nachum2018data}, HRAC \cite{hrac}) achieve temporal abstraction through subgoal decomposition, effectively addressing long-horizon challenges. However, these methods inherently lack safety guarantees, either assuming intrinsically safe subgoals or relying on carefully engineered reward functions. When integrated with Safe RL, hierarchical decomposition can also simplify the dual optimization problem: the low-level policy can focus on optimizing short-horizon objectives while the high-level policy operates over long-horizon subgoals, enabling more stable and efficient learning under safety constraints.

\textbf{Safe HRL Approaches}. The integration of safety constraints with hierarchical RL remains largely unexplored. Existing solutions exhibit critical limitations: Safe HiRO \cite{safetylayerhiro} depends on differentiable cost functions, LyapunovRRT \cite{lyapunovrrt} requires full state observability for region of attraction computation, and IAHRL \cite{iahrl} is restricted to specific domains through handcrafted options. Our proposed ITES framework overcomes these limitations by operating under partial observability while handling unknown, non-differentiable cost functions, and providing verifiable subgoal safety without domain-specific constraints. To our knowledge, ITES represents the first comprehensive solution combining model-based HRL with safety-constrained optimization under these general conditions.

\section{Preliminaries}\label{sec:preliminaries}

\subsection{Constrained Goal-Conditioned MDP}

We investigate the mathematical formulation of the Safe Exploration problem introduced in \cite{safetygym} through the framework of CMDP, represented as $\langle \sS, \sA, \sG, p, R, c, d, \mu, \gamma \rangle$. The environment transition function $p(s'|s, a)$ specifies the probability density of reaching $s' \in \sS$ after taking action $a \in \sA$ in state $s \in \sS$. The initial state distribution $\mu(s_0)$ defines the probability density of beginning an episode at state $s_0$. The variable $g\in\sG$ denotes the final agent goal in the goal space $\sG$. It is generally assumed that the agent does not know the transition dynamics $p(s'|s, a)$. For each transition $\langle s, a, s'\rangle$, the environment produces a scalar external reward $r(s, a, s')$ and another scalar $c(s, a)$ as the cost. In our framework, we treat c as a state-dependent function, i.e., $c(s, a)=c(s)$. The parameter $d \in \sR$ represents the cost limit, indicating the maximum allowable sum of costs over an episode. The optimization problem is to maximize cumulative external rewards from an environment: $J(\pi)=\E_{\pi}\sum_t \gamma^t r(s,a,s')$, while satisfying cost constraint: $J_c(\pi) \leq d$.

Following \cite{hrac}, we employ a hierarchical framework to solve the problem. We adopt a hierarchical agent policy $\pi$ with two levels: a high-level controller with policy $\pi^h_{\theta_{h}}(s_g|s, g)$ and a low-level controller with policy $\pi^l_{\theta_{l}}(a|s, s_g)$ with $s_g\in \sG$ denoting the subgoal generated by policy $\pi^h_{\theta_{h}}$. These controllers are parameterized by separate neural network approximators, with parameters $\theta_{h}$ and $\theta_{l}$, respectively. The high-level controller seeks to maximize the external return:
\begin{equation}
    J_{ex} = \E_{\pi^h_{\theta_{h}}}\sum_i{\gamma^i r^h_{i}}, \quad
    r^h_{i}=\sum_{t=ki}^{t=ki+k-1} r(s_t, a_t, s_{t+1})
\end{equation} 
This policy generates high-level actions in the form of subgoals $s_g \sim \pi^h_{\theta}(s_g|s, g)$, where $s_g, g \in \sG$, at intervals of $k$ time steps ($k > 1$ is a predefined hyperparameter). The goal space $\sG$ is a sub-space of $\sS$, with a known mapping function $\phi: \sS\rightarrow \sG$. The low-level policy performs a primary action $a \sim \pi^l_{\theta_l}(a|s, s_g), a \in \sA$ at every time step. This policy is modulated by intrinsic rewards $r_{in}$ for reaching subgoals generated by high-level controller. The intrinsic reward $r_{in}(s,s_g) = - ||\phi(s) - s_g||$ is a negative Euclidean distance between mappings of current state $s$ and subgoal $s_g$. The low-level objective is to maximize the cumulative intrinsic rewards $J_{in}$:
\begin{equation}
    J_{in}(\pi^l_{\theta_{l}}) = \E_{s_g \sim \pi^h_{\theta_{h}}, \pi^l_{\theta_{l}}} \sum_{i=t}^{i=t+k} r_{in}(s_i, s_g) 
\end{equation} 

The overall objective for the agent policy is to find such parameters $\theta_l, \theta_h$ that the $J_{ex}$ and $J_{in}$ are maximized and the cumulative cost $J_c(\pi) = \E_{\pi^h_{\theta_h}, \pi^l_{\theta_l}} \sum_{\tau} \sum_{t=k\tau}^{k\tau+k-1} c(s_t, a_t)$ is limited:
\begin{equation}
\begin{aligned}
    &\pi^*_{\theta_{h}} = \arg\max J_{ex}(\pi^h_{\theta_{h}}), \quad
    \pi^*_{\theta_{l}} = \arg\max J_{in}(\pi^l_{\theta_{l}}) \phantom{\pi^*_{\theta_{h}} = \arg\max J_{ex}(\pi^h_{\theta_{h}}), \quad}
    \text{s.t. } J_c(\pi) \leq d
\end{aligned}
\end{equation}

\section{Method}\label{sec:method}

The proposed approach ensures safety at two complementary levels: \textit{safe subgoal generation} and \textit{safe subgoal execution}. To generate subgoals that lie within the safe region, we employ a cost model $C_{M}$ that classifies states as safe or unsafe, together with the high-level cost value function $Q^h_c(s, s_g, g)$, which estimates the expected cumulative cost of selecting a subgoal $s_g$. These components guide the high-level policy toward subgoals that are both safe and feasible.

To ensure that the agent can safely reach these subgoals, we incorporate imagined safety through a world model $M$, which predicts short-horizon transitions $s' = M_{\theta_m}(s, a)$. Combined with the cost model $C_{M}$, the world model enables safety evaluation of low-level trajectories in imagination, allowing the low-level controller to avoid unsafe behaviors while pursuing its assigned subgoal.

\subsection{Hierarchy Structure}
As a base algorithm for our approach, we adopted HRAC\cite{hrac}  (Hierarchical Reinforcement learning with k-step Adjacency
Constraint), which consists of two policies: high-level controller $\pi^h_{\theta_{h}}(s_g|s, g)$ and low-level controller $\pi^l_{\theta_{l}}(a|s, s_g)$. The main idea is to generate subgoals that are at a specified distance $k$ from the agent's current position. For this purpose, an Adjacency Network $\psi$ is employed, which maps each state $s$ to its embedding $\psi(s)$. This network is trained using the following loss function:
\begin{equation}
    \label{eq:adj_network_update}
    L_{adj} = \E_{s_i, s_j \in S}\;
     l \cdot \max(|| \psi(s_i) - \psi(s_j)|| - \epsilon, 0) + (1 - l) \cdot \max(\epsilon + \delta - || \psi(s_i) - \psi(s_j)||, 0)
\end{equation}

where $\delta > 0$ is a margin between embeddings, $\epsilon$ is a scaling factor, and $l\in\{0, 1\}$ represents the label indicating $k$-step adjacency. This loss function enables the Adjacency Network to predict embeddings for states $s_i$ and $s_j$ that are $k$ steps apart, such that the condition $||\psi(s_i) - \psi(s_j)|| < \epsilon$ is satisfied.

Subsequently, this network is utilized to update the high-level controller based on the TD3 algorithm \cite{td3}, incorporating a component into its loss function:
\begin{equation}
\label{eq:manager_loss}
    L^h = -Q^h_{\theta_{qh}}(s, s_g, g) + \beta^h_{adj} L_{adj}(s, s_g),
\end{equation}

here $\beta^h_{adj}$ is adjacency loss coefficient---the scaling hyperparameter. For the low-level controller, the TD3 algorithm is employed without modifications with the loss function:
\begin{equation}
    L^l = -Q^l_{\theta_{ql}}(s, a, s_g),
\end{equation}

where $Q^h_{\theta_{qh}}$ and $Q^l_{\theta_{ql}}$ are action value functions (approximated by neural networks) which estimate discounted external and internal returns respectively. 
HRAC enables solving long-horizon tasks but cannot be directly applied in CMDP settings. Below we demonstrate how to ensure safety both in subgoal generation and in low-level policy execution.

\subsection{High-level Safety - safe subgoal generation}

\begin{figure}[H]
\includegraphics[width=0.8\textwidth]{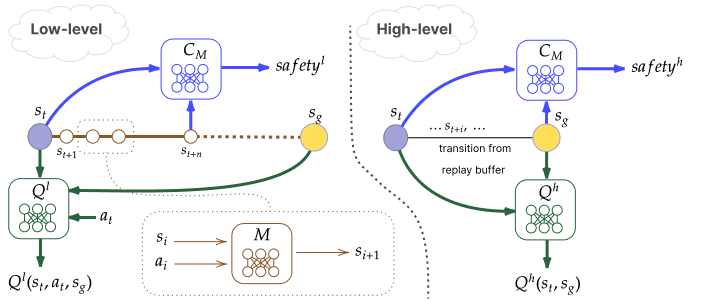}
\caption{The scheme illustrates the training process for the low-level policy and the high-level policy. \textbf{Left}: the cost model $C_{M}$ and the world model $M$ are responsible for ensuring safety during the updates of the low-level policy, while the $Q^l_{\theta_{ql}}$-function is utilized to optimize the reward. \textbf{Right:} the cost model $C_{M}$ is employed for safety considerations in updating the high-level policy, with the $Q^h_{}$-function used for reward optimization.\label{fig:learning_scheme}}
\end{figure}

The scheme for updating the high-level policy is illustrated in Figure~\ref{fig:learning_scheme}. The update is carried out through transitions $\langle s, s_g, R^h \rangle$ and modules such as the cost model $C_M$ and the $Q^h_{\theta_{qh}}$ function, which calculate the safety and utility of the generated subgoal.

\textbf{Cost model.} The objective of the cost model $C_M$ is to predict the probability that an arbitrary subgoal is safe $s_g$ using local state information $s$:
\begin{equation}
    safety^h(s_g, s) = C_{M}(s_g, s) \in[0, 1].
\end{equation}

Since in our experiments the cost function $c$ is binary and is a function of the state $c: \sS \rightarrow \{0,1\}$, the cost model can be used to predict the cost values of states. However, we propose to consider the safety of subgoals as the probability of their safety. Optimizing such safety helps generate goals that are further away from the boundary of safe and unsafe states. For the implementation of the cost model, we utilize a neural network (MLP approximator) parameterized by $\theta_{c}$. We use Binary Cross Entropy Loss as the loss function for learning this MLP. So, the output value of cost model can be considered as
the probability that the state is safe. We train the model in an online manner using the same replay buffer employed for the World Model $M$.

The cost model relies solely on local obstacle observations, yet remains accurate for nearby subgoal prediction as HRAC generates short-horizon subgoals $s_g$ within observable ranges.

\textbf{High-level policy objective.} To ensure that high-level policy generates a safe subgoal, a component $L_{safety^h}(s, s_g)=-safety^h(s_g, s)$ was incorporated into HRAC loss function ({\ref{eq:manager_loss}}) with scaling factor $\beta^h_{safe}$ (safety hyperparameter):
\begin{equation}
\label{eq:safe_manager}
\begin{aligned}
L^h ={}& -Q^h_{\theta_{qh}}(s, s_g, g)
        + \beta^h_{adj} L_{adj}(s, s_g) + \beta^h_{safe} L_{safety^h}(s, s_g)
\end{aligned}
\end{equation}

where $\beta$ factors are predefined hyperparameters that are chosen empirically.

Our experiments revealed that relying solely on the cost model $C_{M}$ to generate safe subgoals is insufficient. In such cases, the high-level policy may learn to predict subgoals within safe regions that are nevertheless infeasible for the low-level policy to reach (e.g., predicting a subgoal behind an obstacle, thereby forcing the low-level policy to perform long-horizon obstacle avoidance). To address this issue, we incorporate the high-level cost value function $Q^h_c(s, s_g, g)$, which estimates the expected discounted cost-to-go until the end of the episode, as an additional objective term for the high-level policy, and learn the Lagrangian multiplier $\lambda$ adaptively to enforce the safety constraint $Q^h_c(s, s_g, g) \leq d$. The resulting high-level objective is:
\begin{equation}
    \label{eq:high_level_policy_update}
    L^h = -Q^h_{\theta_{qh}}(s, s_g, g) + \beta^h_{adj} L_{adj}(s, s_g) + \beta^h_{safe} L_{safety^h}(s, s_g) + \lambda Q^h_c(s, s_g, g)
\end{equation}
The Lagrange multiplier $\lambda$ is updated via dual gradient ascent and projected to $\lambda \ge 0$:
\begin{equation}
\lambda \leftarrow \Big[\lambda + \big(Q_c^h(s,s_g,g) - d\big)\Big]_+ .
\end{equation}

The high-level cost critic $Q_c^h(s, s_g, g)$ is trained by standard Bellman regression on high-level transitions, analogously to the reward critic, using the cumulative cost collected during the corresponding $k$-step subgoal interval.

\subsection{Low-level Safety - safe subgoal execution}

The cost model is used to evaluate subgoals generated by the high-level policy and to optimize them to be safe subgoals. However, while the agent attempts to achieve these safe subgoals, it may violate safety constraints (for example, a mobile robot accelerating excessively while trying to reach a subgoal). To address this issue, we employ a combination of the cost model and the world model (the left learning scheme in Figure~\ref{fig:learning_scheme}).

Given a current subgoal for the agent, $s_g$, we check for safety in imagination by sequentially generating actions using the low-level policy $a \sim \pi^l_{\theta_{l}}(a|s, s_g)$ and then employing the world model to obtain the next state $s' = M(s, a)$, repeating this procedure $n$ times. Here, $n$ is the number of imagination steps that is chosen from discrete uniform distribution $n \sim U(\{1..k-1\})$ (the left part of Figure~\ref{fig:learning_scheme}). The total safety for low-level controller is calculated using the cost and the world models as follows:
\begin{equation}
    safety^l(s, s_g) = C_{M}(\phi(s_n), s) \in[0, 1].
\end{equation}
Although we evaluate safety at a single randomly sampled imagined step, sampling $n$ uniformly makes this an unbiased estimate of average rollout safety, so minimizing it improves safety along the whole trajectory in expectation. Employing cumulative imagined safety costs, instead of random sampling $n$, demonstrates inferior training stability and worse safety performance. We attribute this to two key factors: (1) Computing cost values via the cost model for all imagined states leads to gradient explosion, particularly with long imagination horizons (10-20 steps), and (2) The well-known world model limitation where prediction errors accumulate over extended imagination horizons. By using random states for safety evaluation instead of full rollouts, we avoid lengthy trajectory computations while simultaneously mitigating accumulated prediction errors - an approach that also decreases time for training.

The resulting value of $safety^l(s, s_g)$ is then utilized in the total actor loss:
\begin{equation}
    \label{eq:low_level_policy_update}
    L^l = -Q^l_{\theta_{ql}}(s, a, s_g) + \beta^l_{safe} L_{safety^l}(s, s_g),
    \text{ where } L_{safety^l}(s, s_g) = -safety^l(s, s_g)
\end{equation}

here $\beta^l_{safe}$ is a hyperparameter similar to $\beta^h_{safe}$ from ({\ref{eq:safe_manager}}). Imaginary safety can be added to the high-level policy as an additional loss rather than to the low-level one. However, since predicting subgoals is a more complex task, and based on our experiments, the low-level policy learns to achieve subgoals more quickly than the high-level policy, we incorporate imaginary safety into the low-level policy. To train the cost model and the world model, we utilize a warm start by executing $30,000$ random steps in the environment and pretrain both models over $100$ epochs.

\subsection{Practical Implementation}\label{sec:practical_implementation}

The ITES approach is summarized in Algorithm~\ref{algo:ites}, which details the training loop for both the high-level and low-level policies. Two key auxiliary components are employed: the cost model $C_{M_{\theta_{c}}}$ and the world model $M$.

For the cost model, a replay buffer is maintained with a balanced distribution of safe and unsafe states. This strategy mitigates class imbalance and reduces the risk of overfitting, a common challenge in Safe RL scenarios.

For the world model, we adopt a lightweight predictive model from \cite{mbpo} rather than more complex architectures such as RSSM (which is used in Dreamer \cite{dreamer}), which typically require extensive training. Since our method operates with short-horizon subgoals, a short-horizon world model suffices, reducing cumulative prediction errors and enabling efficient planning within each subgoal horizon. Specifically, we model the transition function $p(s'|s, a)$ using an ensemble of $n$ neural networks, denoted as $M_{\theta_m}$. Each model $M_i$ predicts the next state $s' = M_i(s,a)$, and the final prediction is obtained by averaging over the ensemble, $M_{\theta_m}(s, a) = \frac{1}{n}\sum_{i = 1}^{n} M_i(s,a)$, which helps reduce both epistemic and aleatoric uncertainties.

\begin{algorithm}[H]
   \caption{ITES}
   \label{algo:ites}
\begin{algorithmic}
   \STATE {\bfseries Input:} Environment $env$, High-level policy $\pi^h$, low-level policy $\pi^l$, world model $M$, cost model $C_M$, adjacency network $\psi$, high-level action frequency $k$, number of training episodes $N$, adjacency learning frequency $L$.
   \STATE Pretrain $M_{\theta_m}$ and $C_M$ using random experience.

   \FOR {$n=1$ {\bfseries to} $N$}
       \STATE Reset the environment and sample the initial state $s_0$.
       \STATE $t = 0$.
       \REPEAT
       \IF {$t\equiv 0\,(\mathrm{mod}\ k)$}
         \STATE Sample subgoal $s_{g} \sim \pi^h_{\theta_{h}}(s_g|s_t, g)$.
       \ENDIF
       \STATE Sample low-level action $a_t \sim \pi^l_{\theta_l}(a_t|s_t,\,s_{g})$.
       \STATE Make step $s_{t+1}, r_t, c_t, done = env(s_t, a_t)$.
       \STATE Calculate reward $r_{in} = r_{in}(s_{t+1},s_{g})$.
       \STATE $t = t+1$.
     \UNTIL {$done$ is $true$.}
     \STATE Train $C_M$ using a class-balanced replay buffer.
     \STATE Train world model $M_{\theta_m}$.
     \STATE Train high-level policy $\pi^h$ using Eq.~\ref{eq:high_level_policy_update}.
     \STATE Train low-level policy $\pi^l$ using Eq.~\ref{eq:low_level_policy_update}.
     \IF {$n\equiv 0\,(\mathrm{mod}\ L)$} 
         \STATE Fine-tune $\psi$ using Eq.~\ref{eq:adj_network_update}.
     \ENDIF
   \ENDFOR
\end{algorithmic}
\end{algorithm}

\section{Experiments}\label{sec:experiments}

\begin{table}[H]
\caption{\textbf{Final performance on SafeAntMaze and SafePusher.} Where SAM-C - SafeAntMaze Cshape, SAM-W - SafeAntMaze Wshape, SR - success rate. Methods satisfying the safety condition (Final Cost $\leq d$) are highlighted in \textbf{green}, while non-compliant methods are marked in \textbf{red}. Results are averaged over 5 random seeds}
\resizebox{0.8\linewidth}{!}{
\begin{tabular}{p{0.1\linewidth}p{0.15\linewidth}p{0.19\linewidth}p{0.22\linewidth}}
 Env & Method & Final SR$\uparrow$ & Final Cost$\downarrow$
 \\
 \hline 
 \hline 
  \rowcolor{green!10} & ITES & $\textbf{0.88} \pm 0.01$ & $15.2 \pm 17.3$
  \\
  \rowcolor{green!10}& SafeDreamer & $0$ & $13 \pm 18.11$
  \\
  \rowcolor{green!10} SAM & CUP & 0 & $\textbf{6.8} \pm 22.4$
  \\
  \rowcolor{green!10} C & FOCOPS & 0 & $19.4 \pm 50.9$
  \\
  \rowcolor{red!10} & TD3L/R & $0.38 \pm 0.04$ & $83.6 \pm 37.2$
  \\
  \rowcolor{red!10} & PPOLAG & $ 0.2 \pm 0.2$ & $41.2 \pm 1.1$
  \\
  \rowcolor{red!10} & TD3LAG & $0.13 \pm 0.33$ & $133.45 \pm 184.4$
  \\
  \hline 
  \rowcolor{green!10} & ITES & $\textbf{0.35} \pm 0.05$ & $183 \pm 16.5$
  \\
  \rowcolor{green!10} & TD3L/R & $0.05 \pm 0.02$ & $154 \pm 13.5$
  \\
  \rowcolor{green!10} SAM & CUP & 0 & $\textbf{103.5} \pm 21.5$
  \\
  \rowcolor{green!10} W & FOCOPS & 0 & $141.0 \pm 26.0$
  \\
  \rowcolor{green!10} & TD3LAG & $0$ & $128.68 \pm 210$
  \\
  \rowcolor{green!10} & PPOLAG & $0$ & $105 \pm 20.1$
  \\
   \rowcolor{red!10} & SafeDreamer & $0$ & $216.3 \pm 199$
  \\
  \hline
  \rowcolor{green!10} & ITES & $\textbf{0.42} \pm 0.02$ & $\textbf{12.8} \pm 0.87$
  \\
  \rowcolor{green!10}& SafeDreamer & $0$ & $6.7 \pm 2.7$
  \\
  \rowcolor{red!10} & TD3LAG & $0.05 \pm 0.03$ & $54.72 \pm 3.16$
  \\
  \rowcolor{red!10} Pusher & CUP & $0.01 $ & $46.14 \pm 0.42$
  \\
  \rowcolor{red!10} & FOCOPS & $0 $ & $55.83 \pm 1.67$
  \\
  \rowcolor{red!10} & PPOLAG & $0$ & $51.25 \pm 1.25$
  \\
  \hline
  \hline
\end{tabular}
}
\label{tab:table_compartion_ant}
\end{table}

In this section, we address four research questions: \textbf{RQ1}—can safe flat-policy methods solve long-horizon tasks while satisfying constraints? \textbf{RQ2}—can ITES solve long-horizon tasks under the prescribed cost limit? \textbf{RQ3}—how does ITES perform on short-horizon SafetyGym tasks where a flat policy is sufficient? \textbf{RQ4}—how do the proposed components (high-level safe subgoal generation and low-level imagined safety) contribute to overall safety and performance?

To address these research questions, we developed two long-horizon benchmarks — SafeAntMaze and SafePusher — based on the Gymnasium environments AntMaze and Pusher. We trained flat policy approaches (FOCOPS, CUP, TD3LAG) to evaluate RQ1 and compared them against our proposed ITES method on the long-horizon benchmark. Due to the lack of other fully trainable Safe RL+HRL methods in the literature, we additionally developed HRAC-LAG, a hierarchical baseline, for RQ2. Furthermore, we evaluated our approach on the short-horizon SafetyGym benchmark against additional flat policy baselines (PPOLAG) and a model-based method (MBPPOL). All experiments were run on an NVIDIA GeForce RTX 3060 (8GB). For the same number of environment steps, ITES required 24h (AntMaze) and 12h (SafetyGym), while SafeDreamer required 72h and 34h, respectively, due to a heavier world model.

\subsection{Environments Description}

\textbf{Long-Horizon environments.} SafeAntMaze is a safety-wrapped variant of Mujoco AntMaze, where an 8-DoF ant navigates with partial observability (coordinates, joint angles, and velocities; $\sS \subset \R^{30}$). Positions within distance $dist$ from walls incur a cost (+1), with $d=25$ (C-shaped) and $d=200$ (W-shaped) thresholds. Episodes terminate at 500 steps. SafePusher adapts the Mujoco Pusher task: a 7-DoF arm must push an object to a target using torque control ($\sA \subset \R^7$). Observations ($\sS \subset \R^{24}$) include joint, object, and target states. The task is long-horizon due to high-dimensional control complexity, with a cost limit $d=25$ and 100-step episodes.

We classify these two benchmarks as long-horizon due to the high-dimensional action space of the agent and the correspondingly large state vector representing the robot's configuration
(see Appendix~\ref{app:long_envs}).

\textbf{Short-Horizon environments.} 
In the tasks from the Safety Gym benchmark \cite{safetygym}, the optimal policy achieves the goal within 100 to 200 steps on average. We classify this benchmark as short-horizon due to the robots' simple dynamics and straightforward navigation task. By default, the environment utilizes the Euclidean distance as the reward function, while the cost function assigns a value of +1 when the agent is located within a hazardous zone. Additionally, we consider the PointGoal1 environment from SafetyGym, with sparse reward structure, granting a reward of +1 for reaching the goal. We refer to this environment as PointGoal sparse. Training in an environment with sparse rewards is generally considered to be a more challenging task. We use cost limit $d=40$ for this environment. See Appendix~\ref{app:short_environments} for additional details on the short-horizon environments.

\subsection{Metrics}
Following \cite{safetygym}, we report (i) performance as the \textbf{average success rate} (or episodic reward) over $E{=}40$ evaluation episodes, (ii) the \textbf{average undiscounted episodic cost return} $\hat{J}_c=\frac{1}{E}\sum_{i=1}^{E}\sum_{t=0}^{T_{\mathrm{ep}}} c_t$, and (iii) the \textbf{cost rate} $\rho=\frac{1}{T}\sum_{t=0}^{T} c_t$, where $T$ is the total number of environment interaction steps.

\subsection{Baselines}
We consider two types of baselines: hierarchical (HRAC, HRAC-LAG) and flat-policy-based: CUP \cite{cup}, FOCOPS \cite{focops}, MBPPOL \cite{mbppol}, TD3LAG \cite{safetygym}. To obtain a fair comparison of our safety component to HRAC algorithm against the Lagrangian-based approach, we added a Lagrange multiplier to the high-level policy of the HRAC algorithm and named this as \textbf{HRAC - LAG} that serves as a Safe RL + HRL baseline. To compare with model-based approaches, we propose the \textbf{MBPPOL} algorithm, an algorithm consisting of a single policy that utilizes a world model to train a Lagrangian PPO in a simulated environment and also consider the current state-of-the-art in Safe RL, \textbf{SafeDreamer} \cite{safedreamer}, which employs a world model based on RSSM, a Lagrangian multiplier for constraint handling, and planning in the latent space. To provide a comparison with off-policy approaches, we propose the \textbf{TD3LAG} algorithm, which consists of the TD3 policy and a Lagrangian multiplier. \textbf{TD3L/R} extends TD3LAG with curriculum learning, generating intermediate subgoals upon failure to enhance exploration through progressive target achievement. 

We do not adopt LyapunovRRT, IAHRL, or Safe HIRO as baselines because they are incompatible with the environments proposed in this work due to their inherent limitations. Specifically, LyapunovRRT requires ground-truth obstacle positions, while all considered environments operate under partial observability; IAHRL employs options that are infeasible to implement in these settings; and Safe HIRO necessitates prior knowledge of the cost function and its differentiability, yet agents in our tasks lack access to such information, and the cost functions used are inherently non-differentiable. 
Nevertheless, we performed quantitative comparisons with SafeHIRO (see Appendix~\ref{app:safetybullet}) and qualitative comparisons with IAHRL (see Appendix~\ref{app:other_safe_hrl}).

\subsection{Long-Horizon Task Solving via Safe Flat Policies (RQ1)}
Table \ref{tab:table_compartion_ant} clearly demonstrates the fundamental limitation of flat-policy-based Safe RL methods in solving long-horizon tasks, as evidenced by their near-zero final success rate. The main issue is that these methods fall into a local minimum in reward formulations based on Euclidean distance to the goal: in SafeAntMaze environments the agent must temporarily incur penalties to navigate around obstacles, but flat-policy-based policies greedily maximize immediate reward and therefore fail to reach the target — consistently reflected in their near-zero success rates (SAM C - SafeDreamer: 0, CUP: 0, C-FOCOPS: 0) despite widely varying safety-violation metrics. Although curriculum learning used in TD3L/R helps improve performance, this technique is highly engineering-dependent and cannot be reliably applied across arbitrary environments, limiting its generality. Notably, these methods achieve low safety violation metrics (SAM C - CUP: 6.8, FOCOPS: 19.4, SafeDreamer: 13) primarily because the agent remains stationary and thus avoids violating safety constraints, rather than through successful navigation.

\subsection{Evaluation on long-horizon tasks (RQ2)}
\begin{figure}[H]
    \includegraphics[width=0.7\linewidth]{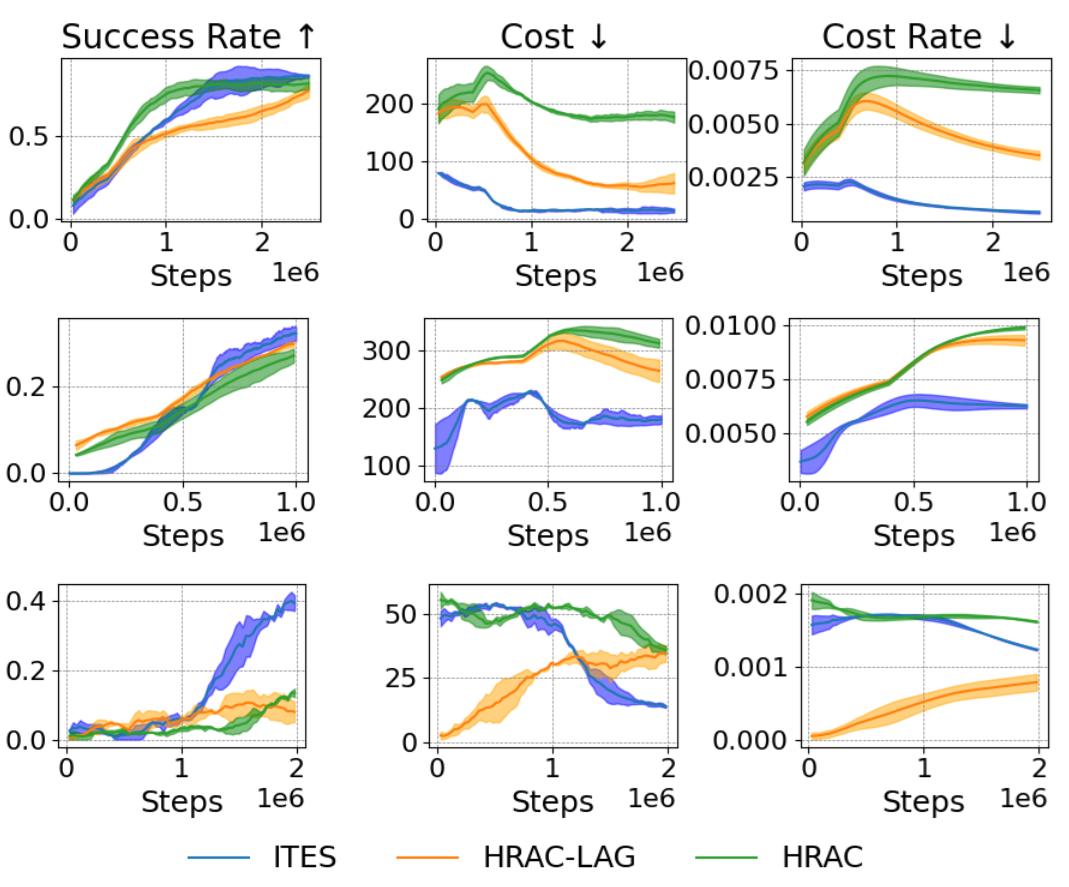}
    \caption{\textbf{Baselines comparison on SafeAntMaze, SafePusher benchmarks}. Comparison of our proposed ITES method with the safe HRAC-LAG and the unsafe HRAC methods. \textbf{First row}: SafeAntMazeCshape environment. \textbf{Second row}: the SafeAntMazeWshape environment. \textbf{Third row}: the SafePusher environment. Each run was conducted with 5 seeds. The shaded area represents the standard deviation.}
\label{fig:baselines_compartion_safe_ant_maze}
\end{figure}

Figure~\ref{fig:baselines_compartion_safe_ant_maze} demonstrates ITES's superior performance versus HRAC-LAG and HRAC across SafeAntMaze and SafePusher benchmarks. While all methods achieve comparable success rates (0.89-0.93) in SafeAntMazeCshape, ITES reduces cost violations by 42\% through its World/Cost Model framework, contrasting with HRAC-LAG's noisy $Q^l_c$ estimates. For SafePusher, ITES improves success rates by 18\% over HRAC, consistent with \cite{nlp_saferl}, as the cost function's dual role as both constraint and reward signal enables more efficient task completion while maintaining safety. The stabilized learning dynamics from model-based safety verification account for these performance advantages.

Across all three long-horizon environments(in Table \ref{tab:table_compartion_ant}), ITES substantially outperforms all other baselines in success rate while remaining safety-compliant. Compared to SafeDreamer, ITES increases success rate by 88\% in SAM-C, 35\% in SAM-W, and 42\% in SafePusher, with an average improvement of ~55\% across the three environments. At the same time, ITES satisfies safety constraints in all settings.

\subsection{Evaluation on short-horizon tasks (RQ3)}
\begin{figure}[H]
\includegraphics[width=0.7\linewidth]{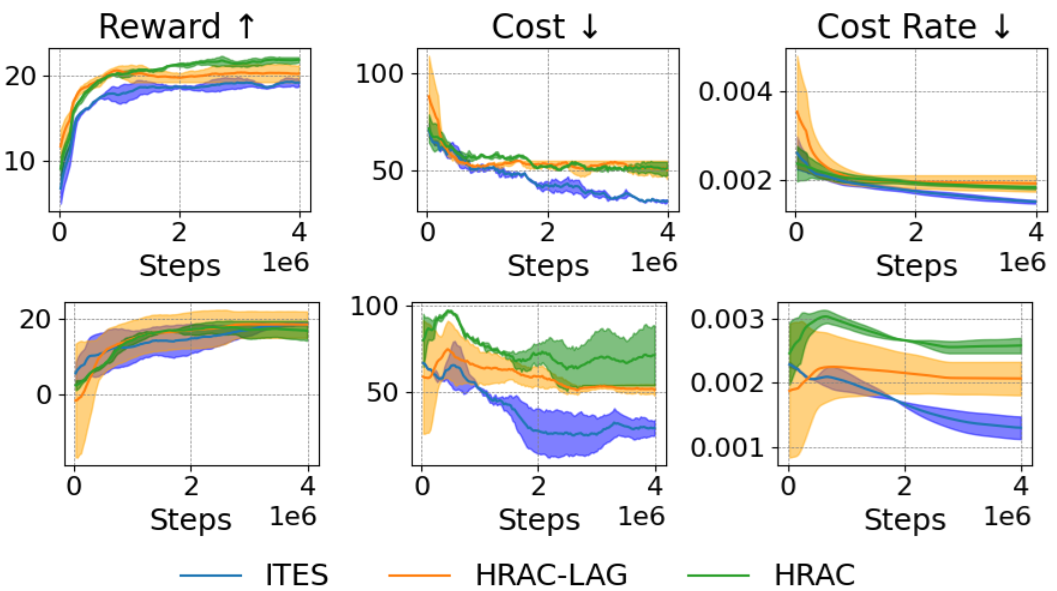}
\caption{\textbf{Baselines comparison on SafetyGym}. \textbf{First row}: PointGoal1. \textbf{Second row}: CarGoal1. Results are averaged over 5 random seeds}
\label{fig:baselines_compartion_safety_gym_point}
\end{figure}

In the experiments on the SafetyGym benchmark, Figure \ref{fig:baselines_compartion_safety_gym_point} presents a comparison with hierarchical policies. It is observed that in the PointGoal1, CarGoal1 environment, our method outperforms HRAC-LAG in both safety while maintaining comparable success rates. 

Table \ref{tab:table_compartion} provides a comparison of the final weights of flat-policy-based approaches (CUP and FOCOPS) with our method ITES on SafetyGym. In the CarGoal1 environment, which features more complex agent dynamics and a larger state space compared to PointGoal1, flat-policy-based methods struggle to maintain high performance: TD3-Lagrangian achieves only 8.21, and SafeDreamer 6.6, as these approaches adopt conservative behaviors and solve fewer tasks. Consequently, they appear safer, achieving lower costs (e.g., SafeDreamer cost = 0), yet such behavior is suboptimal under the task’s safety specification, where a policy is considered safe if its cost $\leq 40$. In contrast, our method successfully balances performance and safety, achieving a higher reward (PointGoal1: $19.6$, CarGoal1: $18.3$) while keeping cost below the threshold.

The results on the PointGoal sparse environment (see Table~\ref{tab:table_compartion}) indicate that simple model-free flat-policy-based algorithms are unable to learn to solve the task, with performance scores of CUP at 1.755, and TD3LAG at 0.07. Such a low success rate indicates that both approaches fail to solve the task effectively. In the case of CUP, the agent remains stationary, resulting in minimal cost, while TD3LAG executes greedy actions but cannot learn to solve the task due to sparse reward conditions. 
The MBPPOL algorithm utilizes ground truth obstacle information (instead of LiDAR data during training) in PointGoal1, PointCar1, and PointGoal sparse environments in the original implementation of the approach. This leads to an unfair comparison, which is why we have marked its metrics with a "-" sign. Nevertheless, even in this setup, the approach performs worse than our method on PointGoal sparse tasks, as it encounters difficulties when learning in sparse reward scenarios. Meanwhile, SafeDreamer achieves a final reward of only 0.5 in the sparse setting, as it employs a reward model trained on an imbalanced dataset due to the sparse rewards; this leads to overly conservative behavior since the method cannot learn to solve tasks requiring a longer horizon than the original PointGoal1 task. These results demonstrate that, under similar safety conditions, ITES exhibits superior performance compared to baselines, achieving a score of 8.64, while maintaining safety constraints.

\begin{table}[H]
\caption{\textbf{Final performance on SafetyGym benchmark.} Where PG1 - PointGoal1, CG1 - CarGoal1, PGs - PointGoal1 sparse. Cost limit for each environment $d=40$. Results are averaged over 5 random seeds.}
\resizebox{0.8\linewidth}{!}{
\begin{tabular}{p{0.1\linewidth}p{0.15\linewidth}p{0.19\linewidth}p{0.19\linewidth}
}
 Env & Method & Final Reward$\uparrow$ & Final Cost$\downarrow$
 \\
 \hline \hline
  & ITES & 
  $\textbf{19.6}\pm 0.46$ & $32.83\pm 1.97$ 
  \\
  & TD3LAG & $15.75 \pm 0.39$ & $56.22 \pm 0.85$ 
  \\
  & FOCOPS 
  & $14.97 \pm 9.01$ & $33.72 \pm 42.24$ 
  \\
  PG1
  & CUP 
  & $14.42 \pm 6.74$ & $19.02 \pm 20.08$ 
  \\
  & PPOLAG
  & $12.96 \pm 6.95$ & $25.80 \pm 34.99$ 
  \\
  & SafeDreamer
  & $12 \pm 4$ & $10.3 \pm 9.5$ 
  \\
  & MBPPOL & - & -  \\
  \hline
  & ITES & $\textbf{18.3}\pm 0.03$ & $21.72 \pm 0.59$
  \\
  & FOCOPS 
  & $15.23 \pm 10.76$ & $31.66 \pm 93.51$
  \\
  & PPOLAG 
  & $13.27\pm 9.26$ & $21.72 \pm 32.06$
  \\
  CG1 & TD3LAG 
  & $8.21 \pm 6.88$ & $53.7 \pm 9.45$
  \\
  & SafeDreamer
  & $6.6 \pm 1.2$ & $0$ 
  \\
  & CUP & $6.14 \pm 6.97$ & $36.12 \pm 89.56$
  \\
  & MBPPOL & - & - 
  \\
  \hline
  & ITES & 
  $\textbf{8.64}\pm 0.38$ & $33.46\pm 1.47$
  \\
  & FOCOPS 
  & $7.635 \pm 0.2$ & $22.9 \pm 1.5$
  \\
  & PPOLAG 
  & $6.85 \pm{0.76}$ & $32.645 \pm{5.9}$
  \\
  & MBPPOL
  & $6.15 \pm 0.15$ & $34.1 \pm 3.6$
  \\
  PGs
  & CUP 
  & $1.755 \pm 0.95$ & $12.925 \pm 5.3$
  \\
  & SafeDreamer
  & $0.5 \pm 0.5$ & $3.22 \pm 2.34$
  \\
  & TD3LAG 
  & $0.07 \pm 0.01$ & $64.7 \pm 1.91$
  \\
  \hline
  \hline
\end{tabular}
}
\label{tab:table_compartion}
\end{table}

\subsection{Ablation Study: Impact of ITES Components on Safety (RQ4)}

\begin{figure}[H]
    \includegraphics[width=0.7\linewidth]{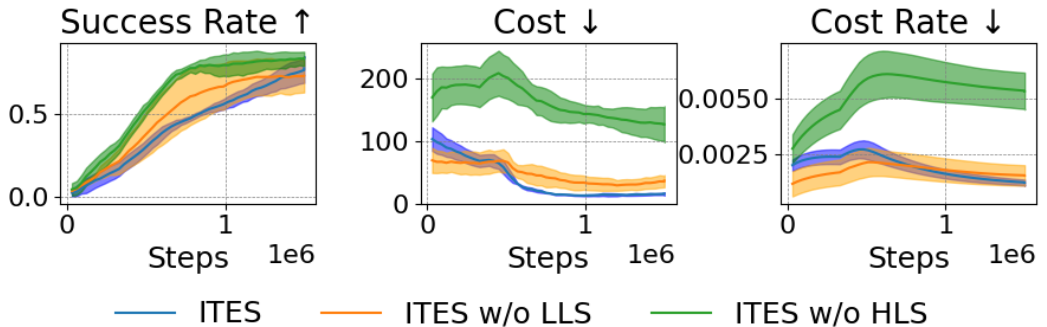}
    \\
    \includegraphics[width=0.7\linewidth]{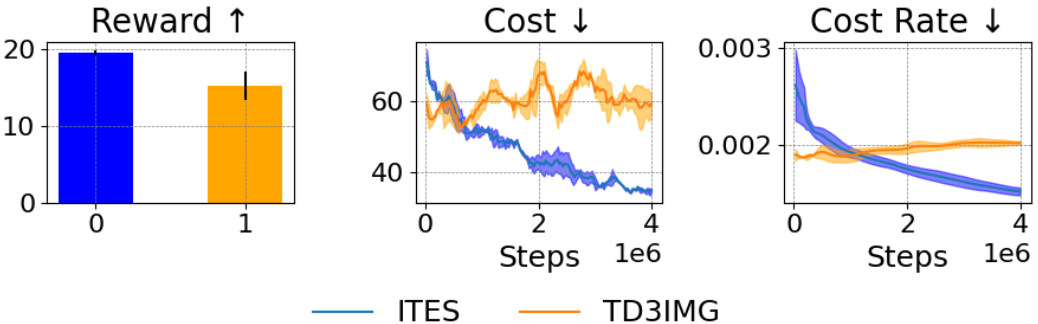}
    \\
    \includegraphics[width=0.7\linewidth]{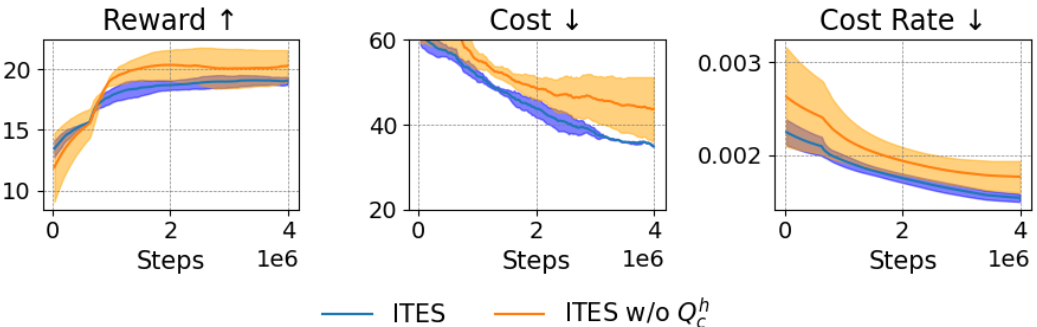}
    
    \caption{\textbf{First row}: Low/High safety level analysis. Different versions of ITES are depicted in the figure, with ITES w/o LLS - ITES without low-level safety implementing safety solely at the high-level and ITES w/o HLS - ITES without high-level safety implementing safety solely at the low-level on the SafeAntMazeCshape. \textbf{Second row}: Comparison of ITES without high-level policy and ITES on the PointGoal1. \textbf{Third row}: Comparison of ITES and ITES without $Q_c^h$ function on the PointGoal1. Plots were obtained across 5 seeds.}
\label{fig:ablation_compartion}
\end{figure}

\textbf{Comparison of ITES w/o HLS and ITES w/o LLS}: To test whether safety can be enforced at only one hierarchy level, we compare ITES to (i) \textbf{ITES w/o LLS} (no low-level safety; safety enforced only at the high level) and (ii) \textbf{ITES w/o HLS} (no high-level safety; safety enforced only at the low level) on \textbf{SafeAntMazeCshape}. As shown in Figure~\ref{fig:ablation_compartion} (first row), high-level safety substantially reduces cost and cost rate, while adding low-level safety further improves safety and reduces variance. In particular, ITES w/o LLS shows higher cost variance because the agent can violate constraints while traveling to a safe subgoal.

\textbf{Comparison of flat ITES (without high-level) and ITES}: The results of training ITES and TD3 with imagined safety computed using the world model, denoted as TD3IMG, are presented in Figure~\ref{fig:ablation_compartion} (second row). Although the cost metric indicates that ITES is significantly safer than TD3IMG, this difference arises because TD3IMG can only minimize safety in imagination over a 10-step horizon (not until the end of the episode); we cannot increase this horizon, as doing so would cause the world model to diverge. In contrast, ITES generates subgoals 10 steps ahead, enabling the TD3 policy to safely reach these goals. 

\textbf{Impact of $Q_c^h$ function on safety performance}: Furthermore, we examined how integrating the Q cost function into the high-level policy's objective loss affects performance, as shown in Figure~\ref{fig:ablation_compartion}, (third row). 

\section{Conclusions}

In summary, we study safe exploration under CMDP constraints and show that flat safe RL baselines struggle on long-horizon tasks. We propose ITES, which combines safe subgoal generation with imagined low-level safety. On long-horizon navigation and manipulation benchmarks, ITES improves success rates while consistently meeting the prescribed safety budget across seeds, and it remains competitive on short-horizon environments.

However, ITES has limitations, including the need to manually design the mapping function $\phi$ from state space to goal space for each task, which restricts its adaptability, particularly to visual input. Future work will focus on addressing these limitations, such as learning goal spaces, improving scalability to enable the application of ITES in real-world robotics tasks.

\vspace{6pt} 

\section*{Author Contributions}
Conceptualization, G.G., A.L. and A.P.; methodology, G.G. and A.L.; software, G.G.; formal analysis, G.G.; investigation, G.G. and A.L.; validation, A.L.; writing---original draft preparation, G.G.; writing---review and editing, G.G., A.L. and A.P.; supervision, A.P. All authors have read and agreed to the published version of the manuscript.

\section*{Funding}
This research was funded by the Ministry of Science and Higher Education of the Russian Federation grant number 075-15-2024-544.

\section*{Data Availability}
The source code supporting the findings of this study is available at \url{https://github.com/Gricha1/ITES_alg}.

\section*{Conflicts of Interest}
The authors declare no conflicts of interest.

\appendix

\section{Cost and World Models Training}
\label{app:cost_world}
\begin{figure}[H]
    \includegraphics[width=0.6\linewidth]{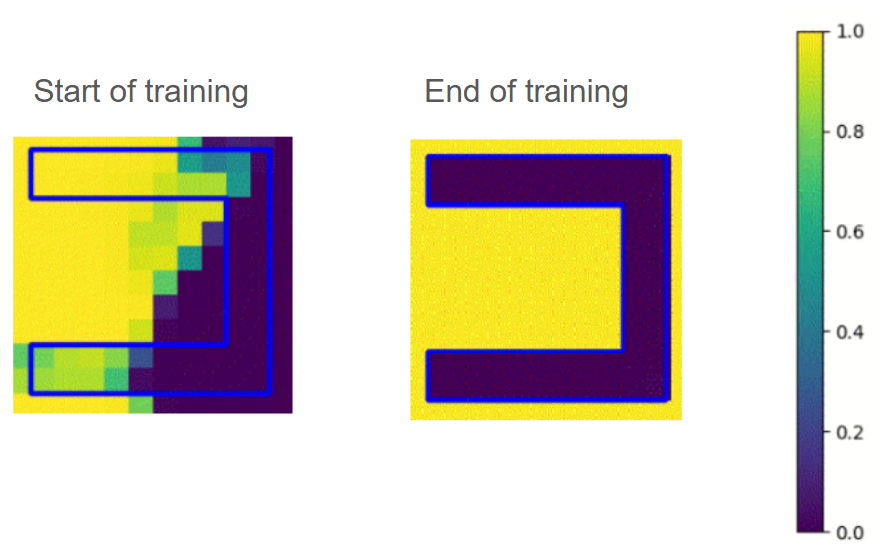}
    \caption{\textbf{Heatmap visualization of the Cost Model in the SafeAntMazeCshape environment}. The blue contour demarcates the safe zone boundaries. \textbf{Left}: Heatmap during early training ($\leq80k$ steps). \textbf{Center}: Heatmap after full training ($1\times2.5^6$ steps). \textbf{Right}: Color bar indicating state safety (1: hazardous, 0: safe).}
\label{fig:cost_model_ablation}
\end{figure}

\begin{figure}[H]
    \includegraphics[width=0.7\linewidth]{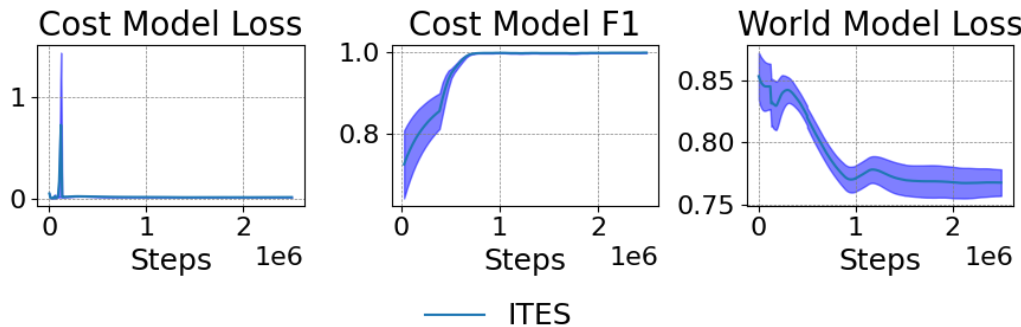}
    \caption{\textbf{Training metrics for the ITES approach's Cost Model and World Model in the SafeAntMazeCshape environment}: Cost Model Loss (Binary Cross-Entropy, BCE), World Model Loss (Mean Squared Error, MSE), and Cost Model F1-score, evaluated on a $30k$-step dataset collected through random policy exploration.}
\label{fig:cost_world_model_losses}
\end{figure}

We further present the training curves for both the cost model and world model components of our approach in the AntMazeCShape environment. Figure \ref{fig:cost_model_ablation} displays the cost model's heatmap visualization, demonstrating that after a $30k$-step warm-start phase, the model already identifies $20-30\%$ of all hazardous states in the environment. Furthermore, the F1-score progression shown in Figure \ref{fig:cost_world_model_losses} indicates that approximately $500k$ training steps are sufficient for the cost model to achieve competent performance.

\section{Long-Horizon Environments}
\label{app:long_envs}
\begin{figure}[H]
    \resizebox{0.2\linewidth}{!}{\includegraphics{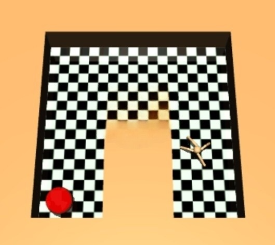}}\\[2mm]
    \resizebox{0.5\linewidth}{!}{\includegraphics{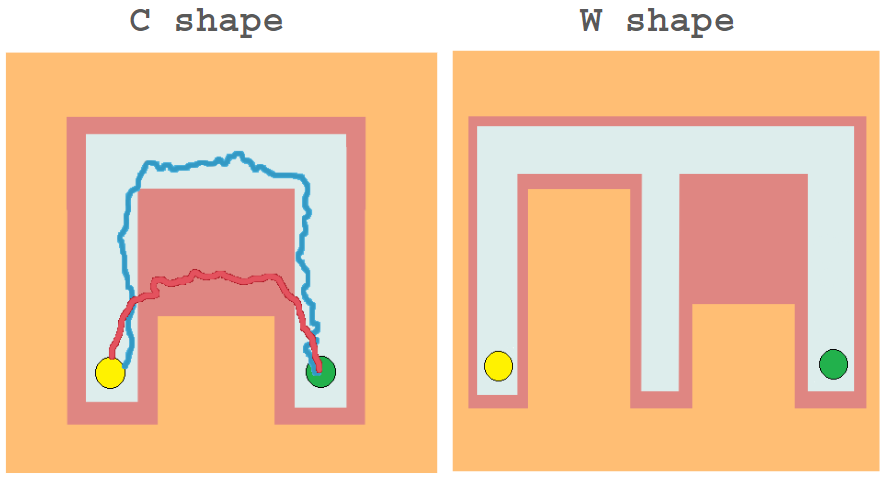}}
    \caption{\textbf{Up}: The figure illustrates the SafeAntMazeCshape environment, where a robot Ant is assigned, and the agent faces a long-horizon task with a specified goal. \textbf{Down}: The scheme of the environments for C shape (left) and W shape (right), where the \textit{green point} represents Ant start pose,  \textit{yellow point} is Ant final goal, the \textit{light field} represents a safe zone, the \textit{pink field} denotes a dangerous area that incurs a cost of $+1$ for each step taken inside it, and the \textit{orange zone} indicates a static obstacle. \textit{The red trajectory} is generated by the HRAC algorithm, which does not take safety into account, while the \textit{blue trajectory} is produced by the ITES algorithm.}
\label{fig:antenv}
\end{figure}

SafeAntMaze (Figure~\ref{fig:antenv}) was created with the safety wrapper for the MujocoAntMaze environment used in the work \cite{hrac}. In SafeAntMaze, the agent is an Ant with action and observation spaces: $\sA \subset \R^8, \sS \subset \R^{30}$. The agent can only observe its current coordinates, joint angles, and angular velocities, lacking information about obstacles. Additionally, we have implemented a safety buffer for the agent's position: if it is within a specified distance, $dist$, from a wall, that position is considered unsafe, resulting in a cost of +1 for the agent, the environment has a maximum episode length of 500 timesteps. We developed two types of maps in this environment: SafeAntMazeCshape and SafeAntMazeWshape. We use $d=25$ and $d=200$ for SafeAntMazeCshape and SafeAntMazeWshape respectively.

\begin{figure}[H]
\includegraphics[width=0.65\linewidth]{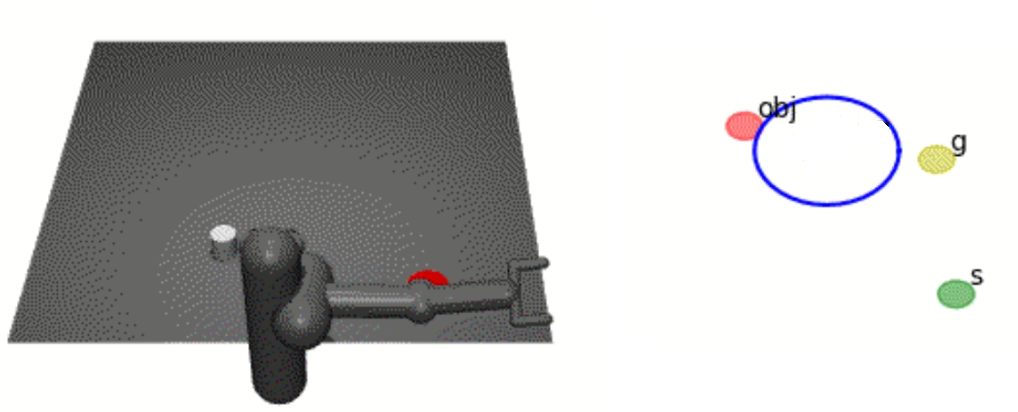}
    \caption{The figure depicts the SafePusher environment. \textbf{Left}: the MuJoCo visualization shows the robotic manipulator in \textit{black}, with the \textit{white object} representing the movable target and the \textit{red marker} indicating its desired final position. \textbf{Right}: displays a top-view schematic of the environment: the \textit{red circle} denotes the manipulable object, the \textit{green circle} represents the end-effector position, the \textit{yellow circle} marks the target location for the object, and the \textit{blue zone} indicates a hazardous area. The agent receives a cost penalty of +1 for every timestep during which the object remains within this hazardous zone.}
\label{fig:pusher_env}
\end{figure}

The SafePusher environment (Figure~\ref{fig:pusher_env}) employs a custom safety wrapper and is built upon the MujocoPusher environment from Gymnasium \cite{brockman2016openai}. In this task, the agent must manipulate a 7-DoF robotic arm to push a cylindrical object to a target position. The observation space consists of a 24-dimensional vector containing information about the arm joints, end-effector position, object position, and target location. The action space is a 7-dimensional vector representing torque values applied to each joint. This constitutes a long-horizon task due to the robot's high-dimensional action space and the inherent complexity of even basic end-effector positioning maneuvers for flat policy approaches, the environment has a maximum episode length of 100 timesteps. We use cost limit $d=25$ for this environment.

\section{Short Horizon Environments}
\label{app:short_environments}
\begin{figure}[H]
    \includegraphics[width=0.3\linewidth]{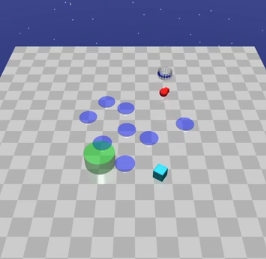}
    \includegraphics[width=0.29\linewidth]{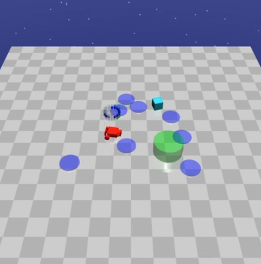}
    \caption{The image presents two tasks: PointGoal1 (left) and CarGoal1 (right) from the SafetyGym environment. \textit{The blue circles} represent hazardous zones, where the agent incurs a cost of +1 while within them. \textit{The green circle} indicates the agent's goal, and the \textit{blue box} represents a movable vase, which the agent can interact with without incurring a penalty.}
\label{fig:gymenv}
\end{figure}

The Safety Gym benchmark \cite{safetygym} establishes a suite of constrained RL tasks requiring agents to accomplish navigation objectives while maintaining safety by avoiding hazardous states. In the PointGoal environment, a point-mass robot with two-dimensional translational and rotational control operates within the workspace(action space: 2), while the CarGoal environment employs a differential-drive car governed by acceleration and steering dynamics(action space: 2). Both environments utilize a unified observation space featuring lidar-like proximity measurements (16 beams) that detect hazards, goal position coordinates, augmented with the agent's kinematic state. The primary task involves navigating to a specified goal position while circumventing hazardous areas. A critical challenge emerges from the agents' inherent kinematic constraints: the point-mass robot's limited maneuverability and the car's non-holonomic dynamics significantly elevate collision risks during exploration, particularly in cluttered configurations. This characteristic makes these environments particularly suitable for evaluating the trade-off between task performance and safety compliance in RL agents.  We use cost limit $d=40$ for this environment. A visual representation of the environment is provided in Figure~\ref{fig:gymenv}.

\begin{figure}[H]
\includegraphics[width=0.3\linewidth]{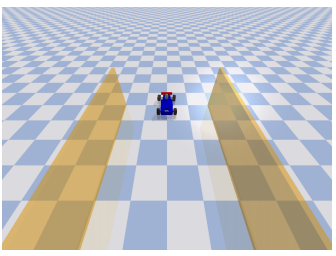}
    \caption{The image illustrates the CarRun environment from the BulletSafetyGym Benchmark. The Car robot is depicted in \textit{blue}, while the boundaries are shown in \textit{orange}. Crossing these boundaries incurs a cost. The agent's objective is to drive straight at the maximum allowable speed; exceeding this speed also results in a cost.}
\label{fig:carrun}
\end{figure}

\begin{table}[H]
\caption{\textbf{Qualitative comparison of ITES, Safe HIRO, and IAHRL.}
Here, Adapt. LL indicates whether a single low-level policy can be reused across tasks, and No diff. cost indicates that the method does not require a differentiable cost function.}
\label{tab:table_ablation_bullet_safety}
\centering
\resizebox{0.5\linewidth}{!}{
\begin{tabular}{lcccc}
\toprule
Method & Adapt.\ LL & No diff.\ cost & Safe SG & WM \\
\midrule
ITES      & Yes & Yes & Yes & Yes \\
Safe HIRO & Yes & No  & No  & No  \\
IAHRL     & No  & Yes & Yes & Yes \\
\bottomrule
\end{tabular}
}
\end{table}

\begin{table}[H]
\caption{\textbf{Final performance on CarRun} comparing ITES and SafeHIRO, where SCRs denotes SafeCarRun with a sparse cost setting and SCRd denotes SafeCarRun with a dense cost setting. The reported safety metric follows the benchmark's original convention and can take negative values because it depends on the agent's speed and distance to the walls.}
\centering
\resizebox{0.7\linewidth}{!}{
\begin{tabular}{p{0.1\linewidth}p{0.13\linewidth}p{0.19\linewidth}p{0.15\linewidth}}
 
 Env & Method & Final Reward$\uparrow$ & Final Cost$\downarrow$
 \\
 \hline 
 \hline
  SCRs & ITES & $\textbf{750.1}$ & $\textbf{-228.7}$
  \\
  & SafeHIRO & $-$ & $-$
  \\
  \hline
  SCRd & ITES & $\textbf{750.1}$ & $-228.7$
  \\
  & SafeHIRO & $216$ & $\textbf{-480}$
  \\
  \hline
  \hline
\end{tabular}
}
\label{tab:table_ablation_safe_hiro}
\end{table}

\begin{table}[H]
\caption{Controller (low-level TD3) hyperparameters. A/P denotes SafeAntMaze and SafePusher, and SG denotes SafetyGym.}
\label{tab:hparams_ctrl}
\centering
\normalsize
\renewcommand{\arraystretch}{1.08}
\begin{tabular}{@{}p{0.4\columnwidth}p{0.15\columnwidth}p{0.15\columnwidth}@{}}
\toprule
\textbf{Hyperparameter} & \textbf{A/P} & \textbf{SG} \\
\midrule
Actor learning rate & 0.0001 & 0.0001 \\
Critic learning rate & 0.001 & 0.001 \\
Replay buffer size & 200,000 & 200,000 \\
Batch size & 128 & 128 \\
Soft update rate & 0.005 & 0.005 \\
Policy update frequency & 1 & 1 \\
$\gamma$ & 0.95 & 0.95 \\
\bottomrule
\end{tabular}
\end{table}

\begin{table}[H]
\caption{Safety-related hyperparameters. A/P denotes SafeAntMaze and SafePusher, and SG denotes SafetyGym.}
\label{tab:hparams_safe}
\centering
\normalsize
\renewcommand{\arraystretch}{1.08}
\begin{tabular}{@{}p{0.54\columnwidth}p{0.15\columnwidth}p{0.15\columnwidth}@{}}
\toprule
\textbf{Hyperparameter} & \textbf{A/P} & \textbf{SG} \\
\midrule
Subgoal safety coefficient $\beta^h_{\mathrm{safe}}$ & 800 & 10 \\
Controller safety coefficient $\beta^l_{\mathrm{safe}}$ & 6 & 0.001 \\
\bottomrule
\end{tabular}
\end{table}

\begin{table}[H]
\caption{Adjacency network hyperparameters. A/P denotes SafeAntMaze and SafePusher, and SG denotes SafetyGym.}
\label{tab:hparams_adj}
\centering
\footnotesize
\setlength{\tabcolsep}{5pt}
\renewcommand{\arraystretch}{1.05}
\resizebox{0.55\columnwidth}{!}{
\begin{tabular}{@{}lcc@{}}
\toprule
\textbf{Hyperparameter} & \textbf{A/P} & \textbf{SG} \\
\midrule
Learning rate & 0.0002 & 0.0002 \\
Batch size & 64 & 64 \\
Online training frequency (steps) & 50,000 (5000) & 50,000 \\
Online training epochs & 25 & 25 \\
Embedding dim & 32 & 32 \\
Hidden dim & 128 & 128 \\
$\epsilon_k$ & 1.0 & 1.0 \\
$\delta$ & 0.2 & 0.2 \\
\bottomrule
\end{tabular}
}
\end{table}

\section{Comparison on the SafetyBullet Benchmark}
\label{app:safetybullet}
Among other hierarchical algorithms addressing safety, SafeHIRO \cite{safetylayerhiro} stands out as the most similar to our approach, as it also employs subgoal generation. The Table~\ref{tab:table_ablation_safe_hiro} provides a comparison between the ITES approach and SafeHIRO in the CarRun environment from the SafetyBullet Benchmark in Figure \ref{fig:carrun}. The CarRun environment features a dense cost function (incorporating information about the distance to walls into the cost); this environment is labeled as cost dense in the table. Since the SafeLayer used in SafeHIRO requires a differentiable cost function for training, SafeHIRO cannot be trained in the original problem formulation where the cost is a binary sparse function. This is why dashes are present in the sparse row for SafeHIRO. It is important to note that the CarRun benchmark uses its own benchmark-specific safety metric. Unlike the nonnegative episodic CMDP cost used in our main experiments, this metric depends on the agent's speed and distance to the walls and can therefore take negative values. 

In contrast, ITES can be trained on both sparse and dense cost functions and achieves consistent results without entering the dangerous zone (where the sparse cost = 0). From the results obtained with a sparse cost function, it is evident that the cost value for SafeHIRO is lower, as it leverages additional information in the cost function specifically, the distance to obstacles encoded as a cost signal while our algorithm does not utilize such information. Nevertheless, our algorithm outperforms SafeHIRO in terms of performance.

Additionally, we provide a table highlighting the differences between the ITES, SafeHIRO, and IAHRL algorithms Table~\ref{tab:table_ablation_bullet_safety}.

\begin{table}[H]
\caption{Cost model and world model hyperparameters. A/P denotes SafeAntMaze and SafePusher, and SG denotes SafetyGym.}
\label{tab:hparams_models}
\centering
\normalsize
\renewcommand{\arraystretch}{1.08}
\begin{tabular}{@{}p{0.42\columnwidth}p{0.15\columnwidth}p{0.15\columnwidth}@{}}
\toprule
\textbf{Hyperparameter} & \textbf{A/P} & \textbf{SG} \\
\midrule
\multicolumn{3}{@{}l}{\textbf{Cost Model}} \\
Initial exploration steps & 30,000 & 30,000 \\
Pretrain epochs & 100 & 100 \\
Batch size & 128 & 512 \\
Buffer size & 1,000,000 & 1,000,000 \\
Learning rate & 0.001 & 0.001 \\
\addlinespace[2pt]
\multicolumn{3}{@{}l}{\textbf{World Model}} \\
Initial exploration steps & 30,000 & 30,000 \\
Pretrain epochs & 100 & 100 \\
Batch size & 256 & 256 \\
Buffer size & 1,000,000 & 1,000,000 \\
Learning rate & 0.001 & 0.001 \\
Train freq & 20 & 20 \\
Num networks & 8 & 8 \\
Num elites & 6 & 6 \\
Hidden size & 200 & 200 \\
\bottomrule
\end{tabular}
\end{table}

\begin{table}[H]
\caption{Manager (high-level TD3) hyperparameters. A/P denotes SafeAntMaze and SafePusher, and SG denotes SafetyGym.}
\label{tab:hparams_mgr}
\centering
\normalsize
\renewcommand{\arraystretch}{1.08}
\begin{tabular}{@{}p{0.42\columnwidth}p{0.15\columnwidth}p{0.15\columnwidth}@{}}
\toprule
\textbf{Hyperparameter} & \textbf{A/P} & \textbf{SG} \\
\midrule
Actor learning rate & 0.0001 & 0.0001 \\
Critic learning rate & 0.001 & 0.001 \\
Replay buffer size & 200,000 & 200,000 \\
Batch size & 128 & 128 \\
Soft update rate & 0.005 & 0.005 \\
Policy update frequency & 10 & 5 \\
$\gamma$ & 0.99 & 0.99 \\
High-level action frequency $k$ & 20 & 10 \\
Reward scaling & 0.1 & 100 \\
Adjacency loss coefficient $\beta^h_{\mathrm{adj}}$ & 20 & 20 \\
\bottomrule
\end{tabular}
\end{table}

\section{Comparison with other Safe RL + HRL Methods}
\label{app:other_safe_hrl}
The Bullet Safety Gym CarRun environment extends the Safety Gym benchmark by implementing a realistic autonomous vehicle navigation task within the PyBullet physics simulator. The environment features a differential-drive car agent controlled through continuous steering and acceleration inputs, operating under non-holonomic constraints that complicate safe navigation. The observation space combines 16-bin LIDAR proximity readings (5m range) for hazard detection, egocentric goal coordinates, and proprioceptive vehicle state including velocity and orientation. The primary task requires navigating to sequentially generated goal positions between static walls, with sparse reward$=+1000$ upon goal achievement and continuous penalty costs ($-1$ per timestep) for hazard proximity. Figure \ref{fig:carrun} illustrates the environment's procedural obstacle generation and the agent's sensor coverage.

\section{Training Details}
\label{app:training}

The hyperparameters are reported in Tables~\ref{tab:hparams_adj}, \ref{tab:hparams_mgr}, \ref{tab:hparams_ctrl}, \ref{tab:hparams_models}, and~\ref{tab:hparams_safe}. Most of these were adopted from HRAC \cite{hrac} and MBPPOL \cite{mbppol}. For the Car and Point environments, we used the same hyperparameters (SG column). Similarly, the hyperparameters for SafeAntMaze-C, SafeAntMaze-W, and SafePusher are reported in the A/P column.

\bibliographystyle{unsrtnat}
\bibliography{nnaabibliography}

@book{cmdpbook,
  title={Constrained Markov Decision Processes},
  author={Altman, Eitan},
  volume={7},
  year={1999},
  publisher={CRC press}
}

@article{brockman2016openai,
  title={Openai gym},
  author={Brockman, Greg and Cheung, Vicki and Pettersson, Ludwig and Schneider, Jonas and Schulman, John and Tang, Jie and Zaremba, Wojciech},
  journal={arXiv preprint arXiv:1606.01540},
  year={2016}
}

@article{nlp_saferl,
  title={Safe reinforcement learning with free-form natural language constraints and pre-trained language models},
  author={Lou, Xingzhou and Zhang, Junge and Wang, Ziyan and Huang, Kaiqi and Du, Yali},
  journal={arXiv preprint arXiv:2401.07553},
  year={2024}
}

@inproceedings{td3,
  title={Addressing function approximation error in actor-critic methods},
  author={Fujimoto, Scott and Hoof, Herke and Meger, David},
  booktitle={International conference on machine learning},
  pages={1587--1596},
  year={2018},
  organization={PMLR}
}

@inproceedings{saclagrangian,
    title={Learning to Walk in the Real World with Minimal Human Effort},
    author={Ha, Sehoon and Xu, Peng and Tan, Zhenyu and Levine, Sergey and Tan, Jie},
    booktitle={Conference on Robot Learning},
    pages={1110--1120},
    year={2021},
    organization={PMLR}
}

@article{dreamer,
    title={Mastering Diverse Domains through World Models},
    author={Hafner, Danijar and Pasukonis, Jurgis and Ba, Jimmy and Lillicrap, Timothy},
    journal={arXiv preprint arXiv:2301.04104},
    year={2023}
}

@inproceedings{safeslac,
    title={Safe reinforcement learning from pixels using a stochastic latent representation},
    author={Hogewind, Yannick and Simao, Thiago D and Kachman, Tal and Jansen, Nils},
    booktitle={The Eleventh International Conference on Learning Representations},
    year={2023}
}

@inproceedings{safedreamer,
    title={SafeDreamer: Safe Reinforcement Learning with World Models},
    author={Weidong Huang and Jiaming Ji and Borong Zhang and Chunhe Xia and Yaodong Yang},
    booktitle={The Twelfth International Conference on Learning Representations},
    year={2024},
}

@inproceedings{huang2021risk,
  title={Risk conditioned neural motion planning},
  author={Huang, Xin and Feng, Meng and Jasour, Ashkan and Rosman, Guy and Williams, Brian},
  booktitle={2021 IEEE/RSJ International Conference on Intelligent Robots and Systems (IROS)},
  pages={9057--9063},
  year={2021},
  organization={IEEE}
}

@inproceedings{mbpo,
    author = {Janner, Michael and Fu, Justin and Zhang, Marvin and Levine, Sergey},
    booktitle = {Advances in Neural Information Processing Systems},
    pages = {},
    publisher = {Curran Associates, Inc.},
    title = {When to Trust Your Model: Model-Based Policy Optimization},
    volume = {32},
    year = {2019}
}

@inproceedings{mbppol,
  title={Model-based safe deep reinforcement learning via a constrained proximal policy optimization algorithm},
  author={Jayant, Ashish K and Bhatnagar, Shalabh},
  booktitle={Advances in Neural Information Processing Systems},
  volume={35},
  pages={24432--24445},
  publisher = {Curran Associates, Inc.},
  year={2022}
}

@inproceedings{
    levy2017hierarchical,
    title={Hierarchical Reinforcement Learning with Hindsight},
    author={Andrew Levy and Robert Platt and Kate Saenko},
    booktitle={International Conference on Learning Representations},
    year={2019},
}

@inproceedings{nachum2018data,
  title={Data-efficient hierarchical reinforcement learning},
  author={Nachum, Ofir and Gu, Shixiang Shane and Lee, Honglak and Levine, Sergey},
  booktitle={Advances in Neural Information Processing Systems},
  volume={31},
  year={2018},
  publisher = {Curran Associates, Inc.},
}

@article{safetygym,
    title={Benchmarking safe exploration in deep reinforcement learning},
    author={Ray, Alex and Achiam, Joshua and Amodei, Dario},
    journal={arXiv preprint arXiv:1910.01708},
    year={2019}
}

@inproceedings{safetylayerhiro,
    title={Safe Robot Navigation Using Constrained Hierarchical Reinforcement Learning},
    author={Roza, Felippe Schmoeller and Rasheed, Hassan and Roscher, Karsten and Ning, Xiangyu and G{\"u}nnemann, Stephan},
    booktitle={2022 21st IEEE International Conference on Machine Learning and Applications (ICMLA)},
    pages={737--742},
    year={2022},
    organization={IEEE}
}

@inproceedings{cup,
  title={Constrained update projection approach to safe policy optimization},
  author={Yang, Long and Ji, Jiaming and Dai, Juntao and Zhang, Linrui and Zhou, Binbin and Li, Pengfei and Yang, Yaodong and Pan, Gang},
  booktitle={Advances in Neural Information Processing Systems},
  volume={35},
  pages={9111--9124},
  year={2022},
  publisher = {Curran Associates, Inc.},
}

@inproceedings{safeditor,
     author = {Yu, Haonan and Xu, Wei and Zhang, Haichao},
     booktitle = {Advances in Neural Information Processing Systems},
     pages = {2608--2621},
     publisher = {Curran Associates, Inc.},
     title = {Towards Safe Reinforcement Learning with a Safety Editor Policy},
     volume = {35},
     year = {2022}
}

@inproceedings{focops,
  title={First order constrained optimization in policy space},
  author={Zhang, Yiming and Vuong, Quan and Ross, Keith},
  booktitle = {Advances in Neural Information Processing Systems},
  volume={33},
  pages={15338--15349},
  publisher = {Curran Associates, Inc.},
  year={2020}
}

@inproceedings{hrac,
    author = {Zhang, Tianren and Guo, Shangqi and Tan, Tian and Hu, Xiaolin and Chen, Feng},
    booktitle = {Advances in Neural Information Processing Systems},
    pages = {21579--21590},
    publisher = {Curran Associates, Inc.},
    title = {Generating Adjacency-Constrained Subgoals in Hierarchical Reinforcement Learning},
    volume = {33},
    year = {2020}
}

@inproceedings{chane2021goal,
  title={Goal-conditioned reinforcement learning with imagined subgoals},
  author={Chane-Sane, Elliot and Schmid, Cordelia and Laptev, Ivan},
  booktitle={International Conference on Machine Learning},
  pages={1430--1440},
  year={2021},
  organization={PMLR}
}

@article{brunke2022safe,
  title={Safe learning in robotics: From learning-based control to safe reinforcement learning},
  author={Brunke, Lukas and Greeff, Melissa and Hall, Adam W and Yuan, Zhaocong and Zhou, Siqi and Panerati, Jacopo and Schoellig, Angela P},
  journal={Annual Review of Control, Robotics, and Autonomous Systems},
  volume={5},
  pages={411--444},
  year={2022}
}

@inproceedings{bassich2019continuous,
  title={Continuous curriculum learning for reinforcement learning},
  author={Bassich, Andrea and Kudenko, Daniel},
  booktitle={Proceedings of the 2nd Scaling-Up Reinforcement Learning (SURL) Workshop. IJCAI},
  year={2019}
}

@article{mandlekar2020learning,
  title={Learning to generalize across long-horizon tasks from human demonstrations},
  author={Mandlekar, Ajay and Xu, Danfei and Mart{\'\i}n-Mart{\'\i}n, Roberto and Savarese, Silvio and Fei-Fei, Li},
  journal={arXiv preprint arXiv:2003.06085},
  year={2020}
}

@article{garcia2015comprehensive,
  title={A comprehensive survey on safe reinforcement learning},
  author={Garc{\i}a, Javier and Fern{\'a}ndez, Fernando},
  journal={Journal of Machine Learning Research},
  volume={16},
  number={1},
  pages={1437--1480},
  year={2015}
}

@inproceedings{lyapunovrrt,
  title={Model-free neural lyapunov control for safe robot navigation},
  author={Xiong, Zikang and Eappen, Joe and Qureshi, Ahmed H and Jagannathan, Suresh},
  booktitle={2022 IEEE/RSJ International Conference on Intelligent Robots and Systems (IROS)},
  pages={5572--5579},
  year={2022},
  organization={IEEE}
}

@article{iahrl,
  title={Imagination-Augmented Hierarchical Reinforcement Learning for Safe and Interactive Autonomous Driving in Urban Environments},
  author={Lee, Sang-Hyun and Jung, Yoonjae and Seo, Seung-Woo},
  journal={IEEE Transactions on Intelligent Transportation Systems},
  year={2024},
  publisher={IEEE}
}
\end{document}